%% file: acl_latex.tex
\pdfoutput=1

\documentclass[11pt]{article}

\usepackage{acl}
\usepackage{amsmath}
\usepackage{afterpage}
\usepackage{algorithm}
\usepackage{algorithmic}
\usepackage{times}
\usepackage{latexsym}
\usepackage{booktabs}
\usepackage{hyperref}
\usepackage{latexsym, graphicx, amsmath, float, tabularx, subcaption}
\usepackage{soul}
\usepackage{subcaption}
\definecolor{lightgreen}{HTML}{c9dfb7}
\definecolor{lightgrey}{HTML}{f2f2f2}
\usepackage{multirow}
\usepackage{makecell}
\usepackage{float}
\usepackage[T1]{fontenc}

\usepackage[utf8]{inputenc}

\usepackage{microtype}

\usepackage{inconsolata}

\usepackage{graphicx}

\title{X-ray Made Simple: Lay Radiology Report Generation and Robust Evaluation}

\newcommand*\samethanks[1][\value{footnote}]{\footnotemark[#1]}

\author{ 
	Kun Zhao\textsuperscript{1}\thanks{\quad \small Equal contribution}\space,
	Chenghao Xiao\textsuperscript{2}\samethanks\space,
	Sixing Yan\textsuperscript{3},
	Haoteng Tang\textsuperscript{5},
	William K. Cheung\textsuperscript{3},
	\\\textbf{Noura Al Moubayed}\textsuperscript{2}, 
	\textbf{Liang Zhan}\textsuperscript{1}\thanks{\quad \small Corresponding authors}\;,
	\textbf{Chenghua Lin}\textsuperscript{\textbf{4}}\samethanks\space\\ 
	\textsuperscript{1}University of Pittsburgh\vspace{-0.5mm} \space
	\textsuperscript{2}Durham University\vspace{-0.5mm} \space
	\textsuperscript{3}Hong Kong Baptist University\vspace{1mm} \\
	\textsuperscript{4}The University of Manchester\vspace{-0.5mm} \\
	\textsuperscript{5}University of Texas Rio Grande Valley\vspace{-0.5mm} \\
	\texttt{
		\{kun.zhao, liang.zhan\}@pitt.edu, 
	}\vspace{-0.5mm} 
	\texttt{
		chenghao.xiao@durham.ac.uk, 
	}\vspace{-0.5mm} \\
	\vspace{-0.5mm}
	\texttt{
		chenghua.lin@manchester.ac.uk
	}\vspace{-0.5mm}
}

\begin{document}
\maketitle
\begin{abstract}
While multimodal generative models have advanced radiology report generation (RRG), challenges remain in making reports accessible to patients and ensuring reliable evaluation.
The technical language and templated nature of professional reports hinder patient comprehension and enable models to artificially boost lexical metrics such as BLEU by reproducing common report patterns.
To address these limitations, we propose the Layman’s RRG framework, which leverages layperson-friendly language to enhance patient accessibility and promote more robust evaluation and report generation by encouraging models to focus on semantic accuracy over rigid templates. 
Our approach also introduces and releases two refined layman-style datasets (at the sentence and report levels), along with a semantics-based evaluation metric that mitigates inflated lexical scores and a layman-guided training strategy.
Experiments show that training on layman-style data helps models better capture the meaning of clinical findings. Notably, we observe a positive scaling law: model performance improves with more layman-style data, in contrast to the inverse trend observed with templated professional language.

\end{abstract}

\section{Introduction}
With the advancement of generative models, image captioning has made significant progress in producing accurate textual descriptions from visual inputs. 
This capability has been increasingly applied in the medical domain, particularly in Radiology Report Generation (RRG) \citep{lin2022sgt, wang2022medclip, lee2023llm, hou2023organ, yan2023style, li2023unify, liu2024bootstrapping}. 
RRG aims to generate descriptive reports from medical images, such as chest X-rays, to reduce radiologists’ workload while improving the quality, consistency, and efficiency of clinical documentation. Despite recent progress, two critical challenges remain underexplored. 
First, the generated reports often lack patient accessibility due to their use of highly technical language and rigid clinical templates, making them difficult for non-experts to understand. 
Second, current evaluation metrics and training paradigms emphasize surface-level textual similarity rather than true semantic understanding, potentially masking important deficiencies in report quality. 
Although these challenges may appear distinct, they are closely linked: the templated language that hinders patient comprehension also leads models to overfit to surface patterns, inflating evaluation scores and hindering semantic generalization.

A patient-centered approach is becoming increasingly vital in modern healthcare, emphasizing transparency and shared decision-making. 
With policies like the 21st Century Cures Act requiring immediate access to electronic health records (EHR), patients now often receive radiology reports before any clinical interpretation. 
However, these reports—designed primarily for clinician communication and billing—are written in highly technical language, with fewer than 4\% meeting the eighth-grade reading level typical of U.S. adults \cite{martin2019readability}. 
This mismatch presents major barriers to understanding and engagement, frequently resulting in confusion, anxiety, and poor adherence to follow-up or treatment plans \cite{domingo2022preventing, mabotuwana2018improving}. 
The challenge is compounded by the fact that only 50\% of recommended follow-ups are completed \cite{mabotuwana2019automated}, in part due to unclear communication of incidental findings. 
While prior studies have explored barriers from the patient’s perspective, little work has addressed the need to redesign the reports themselves. 
Improving report accessibility is therefore both a practical necessity and an ethical obligation in advancing patient-centered AI.

Beyond the challenge of patient accessibility, radiology report generation also faces a fundamental lack of robustness in both evaluation and training. 
On the evaluation side, most RRG models are still assessed using lexical overlap-based metrics like BLEU and ROUGE \citep{papineni2002bleu, lin2004rouge}, which remain dominant in the field \cite{liu2023systematic}. 
However, these metrics operate at the surface level, capturing word-level similarity while ignoring clinical meaning. For example, the phrases “\textit{there is a focal consolidation}” and “\textit{there is no focal consolidation}” receive similarly high BLEU scores due to shared structure, despite expressing opposite clinical conclusions \citep{stent2005evaluating}. 
This shortcoming is magnified by the highly templated nature of radiology reports \citep{li2019knowledge, Kale2023ReplaceAR}, where rigid formats enable models to achieve high scores by mimicking patterns rather than grasping content. 
Prior work has shown that template-based substitutions can produce strong lexical scores even when semantic accuracy is lost \citep{Kale2023ReplaceAR}. 
Moreover, such structural rigidity in professional reports could also effect training, as models exposed to these templates often overfit to superficial cues instead of learning generalizable semantic representations.

We hypothesize that adopting layman-style language in radiology report generation can simultaneously address the dual challenges of accessibility and robustness. 
From the patient’s perspective, layman terms enhance the readability and comprehensibility of reports, making them more inclusive and actionable. 
From the modeling perspective, the linguistic diversity and absence of rigid templates in layman-style reports encourage models to focus on semantic understanding rather than overfitting to superficial patterns. 
Building on this insight, we propose a new framework for radiology report generation grounded in layman’s terms. 
Our framework includes: (1) creating two high-quality \textbf{layman-style datasets}—a sentence-level dataset and a report-level dataset; (2) a \textbf{semantics-based evaluation method} based on layman’s terms, which provides fairer assessments that mitigates inflated BLEU scores; and (3) a \textbf{training strategy based on layman's terms} that improves the model’s semantic learning and reduces its reliance on templated language in professional reports.

To validate the effectiveness of the Layman’s RRG framework, we conduct extensive experiments using the publicly available MIMIC-CXR dataset \citep{johnson2019mimic}. 
Results show that our semantics-based evaluation method, combined with the sentence-level layman dataset, provides significantly more robust assessments. 
Furthermore, models trained with our layman-guided strategy exhibit stronger semantic generalization compared to those trained on templated professional reports. 
Notably, we observe a promising scaling trend: as the amount of layman-style training data increases, model performance continues to improve—unlike the diminishing gains seen with professional report training. 
These findings offer strong empirical support for our hypothesis that layman-style language enhances both accessibility and robustness in radiology report generation.
In summary, our contributions are as follows:
\begin{itemize}
    \item We introduce two high-quality layman-style radiology report generation datasets: a sentence-level dataset and a report-level dataset. 
    To the best of our knowledge, this is the first systematic effort to create patient-friendly datasets for RRG, offering a valuable resource for future research aimed at enhancing the readability and inclusiveness of medical AI systems.    
    \item We propose a layman-guided evaluation method for RRG that leverages LLM-based embedding models to substitute professional report sentences with semantically matched layman equivalents from our dataset. This method enables fairer and more robust assessment using both traditional lexical metrics and our proposed semantics-based metric.

    \item We demonstrate training on our report-level layman dataset enhances the model’s semantic understanding and reveals a promising scaling law: performance improves consistently with more layman-style data—contrasting with the diminishing returns seen when training on professional reports.
\end{itemize}

\section{Related work}
\subsection{Patient-Centric Reports}

Some medical researches show that a direct link between patients’ understanding of their medical information with adherence to recommended prevention and treatment processes, better clinical outcomes, better patient safety within hospitals, and less health care utilization \cite{anhang2014examining, lopez2024automated, martin2019readability}. Radiology reports, although written primarily for healthcare providers, are read increasingly by patients and their family. However, few researches have focused on patient-centric reports.

\subsection{Evaluation Metrics for Radiology Report Generation}
Evaluation metrics are essential for RRG as they provide measurements of the quality of the produced radiology reports from various approaches and ensure a fair comparison among counterparts. Similar to other AI research domains, prevailing approaches in RRG evaluation adopt automatic metrics by comparing the generated reports with gold standard references (i.e., doctor-written reports). Generally, metrics for this task are categorized into five types: natural language generation (NLG) \cite{papineni2002bleu, lin2004rouge, banerjee2005meteor, zhao2023evaluating, zhao-etal-2024-slide, yang2024structured}, clinical efficacy (CE) \cite{peng2018negbio, irvin2019chexpert}, standard image captioning (SIC) \cite{vedantam2015cider}, embedding-based metrics, and task-specific features-based metrics. Among these, NLG metrics and CE metrics are the most widely adopted in current approaches. However, most of these metrics primarily focus on word overlap and do not adequately consider the semantic meaning between the ground truth and generated reports.

\begin{figure*}[tb]
\centering
\includegraphics[width=0.9\linewidth]{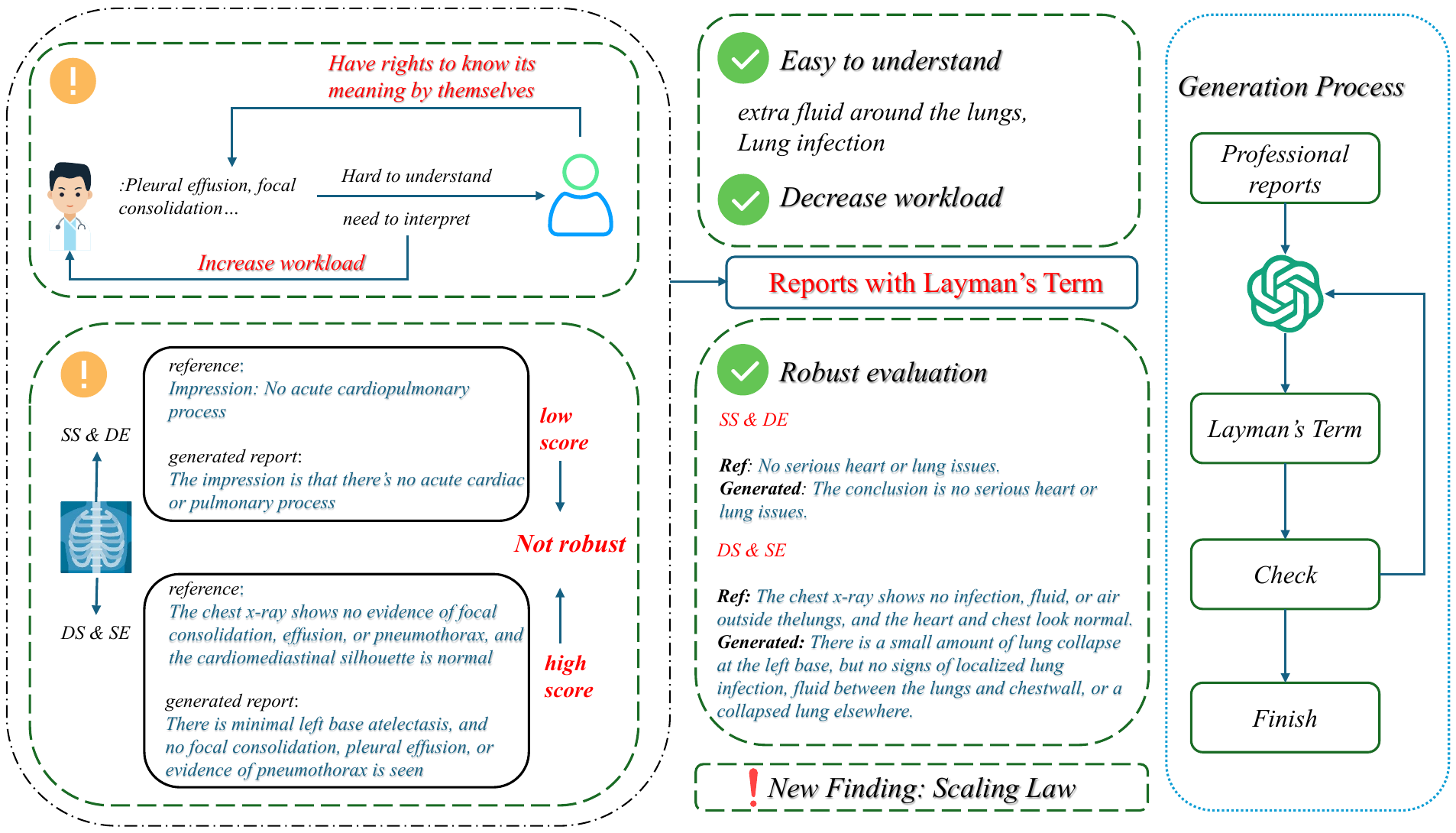}
\caption{The Layman's RRG Framework. The "DS \& SE" denotes different semantics and similar expressions. The "SS \& DE" denotes similar semantics and different expressions.}
\label{fig:model}
\end{figure*}

\section{Layman's Term RRG}
In this section, we present \textbf{Layman’s term RRG}, a unified framework encompassing \{data creation, evaluation, and training\}, designed to address the limitations of lexical-based metrics and the rigid, patterned nature of professional radiology reports.
The framework (see Figure~\ref{fig:model}) is supported by two complementary resources: a sentence-level dataset for semantics-based evaluation and a report-level dataset for training models with improved semantic generalization.

\subsection{Data Creation}
Our data construction pipeline comprises three components: a deduplication preprocessing (applicable only to the sentence-level dataset), a generation–refinement step, and a human verification postprocessing. 
This pipeline is designed to produce high-quality layman-style sentences and reports.

\subsubsection{Deduplication Preprocessing}
We first use NLTK to segment each report into individual sentences. Through analyzing large volumes of reports, we found that many repetitive sentences share similar semantics. To simplify the final dataset and reduce the burden of pairwise similarity computation, we apply extensive deduplication to the sentence-level inputs. To this end, we use GritLM \cite{muennighoff2024generative}, a decoder-based embedding model that achieves state-of-the-art performance on the Massive Text Embedding Benchmark (MTEB) and the Reasoning as Retrieval Benchmark (RAR-b), to encode sentences and obtain their vector representations. We then iteratively compute pairwise cosine similarities between sentences, retaining those that do not exceed a similarity threshold of 0.8 with previously selected sentences and discarding the rest. Through this deduplication procedure, the number of sentences is reduced from approximately 490,000 to 50,000, substantially lowering computational cost and improving the efficiency of subsequent processing.

\subsubsection{Generation–Refinement Step}
\noindent\textbf{Generation.} After the deduplication on sentences, we use \texttt{GPT-4o} to translate professional sentences or reports into layman-style language. The prompt design—detailed in Appendix~\ref{prompt:trans}—specifies the generation objectives, enables batch processing, and instructs the model to return outputs in JSON format.
This approach largely reduces cost and improves output consistency through referencing in-batch examples.

\noindent\textbf{Refinement.}
To enhance translation quality, we introduce a self-refinement method involving a semantic-checking module built upon embedding models, and a correctness self-checking module using the same LLM in the generation step.
Details of the self-check prompt are provided in Appendix~\ref{prompt:refine}.
For each professional–layman sentence pair, we combine self-check feedback from \texttt{GPT-4o} with semantic similarity scores from GritLM to ensure the quality of translated sentence. 
A translation must pass both checks to be accepted; otherwise, the sentence is resubmitted for regeneration. 
The full procedure of the generation-refinement step is outlined in Appendix~\ref{alg:refine}.

\subsubsection{Human Verification}
Following the refinement process, the dataset quality improved substantially. 
As shown in Appendix~\ref{appendix: refinement}, correction rates increase across self-refinement iterations. Additionally, we randomly sampled 500 sentence pairs for human verification, where over 98\% were judged as correct matches.

\subsection{Beyond Lexical Overlap: Semantics-Based Evaluation}
\input{Tables/concept}
Through thorough analysis of radiology reports, we observed that word-overlap metrics such as BLEU, ROUGE, and METEOR do not accurately reflect the quality of generated reports. This discrepancy arises due to the presence of semantically similar sentences with different wordings, as well as semantically different sentences with high lexical overlap. For example, the sentences \textit{``There is a definite focal consolidation, no pneumothorax is appreciated"} and \textit{``There is no focal consolidation, effusion, or pneumothorax"} convey distinct clinical meanings but achieve a BLEU-1 score greater than 0.6. 
This demonstrates that even when the underlying pathology differs, high BLEU scores may still be obtained due to surface-level similarity. 
Conversely, the sentences \textit{``Impression: No acute cardiopulmonary process"} and \textit{``The impression is that there’s no acute cardiac or pulmonary process"} convey the same meaning but receive a low BLEU-1 score due to differences in phrasing.
We categorize these inconsistencies into two types: \textbf{expression difference} issues and \textbf{semantics difference} issues. 
An expression difference issue occurs when the candidate and reference sentences share similar semantics but exhibit low word overlap. 
A semantics difference issue arises when the sentences differ in meaning but have high word overlap. Both issues can result in misleading BLEU scores, as illustrated in Table~\ref{tab:concept}.

To address these issues, we propose a novel evaluation method for assessing generated radiology reports. In brief, the method compares a candidate report with a reference report by first splitting both into individual sentences. Each sentence is then replaced with its most semantically similar counterpart from our constructed sentence-level dataset, using GritLM to compute semantic similarity. Sentences exceeding a predefined similarity threshold are considered matched. We then calculate the proportion of matched sentences in both the candidate and reference reports as an additional metric, reported alongside traditional word-overlap metrics such as BLEU, ROUGE, and METEOR. This complementary metric enables our evaluation framework to mitigate the limitations of lexical-based evaluation and provide a more semantically grounded assessment of report quality. 
The detailed evaluation algorithm is provided in Appendix~\ref{app:eva}.

\subsection{Robust Training with Layman-style Data}
To investigate how training data style affects the semantic generalization ability of generative models, we design a scaling-based training protocol using both professional and layman-style radiology reports. 
Our central hypothesis is that heavily templated professional reports encourage models to focus on surface structure rather than semantic content, while translating these reports into layman’s terms removes rigid formatting and introduces linguistic diversity, thereby promoting semantic learning.

We construct a series of training subsets for both datasets (professional and layman-style), with sizes of 5k, 10k, 15k, 20k, 25k, and 50k samples. 
For each subset, we fine-tune the \texttt{MiniGPT-4} model. 
The training is conducted for 10 epochs with a batch size of 50, using gradient accumulation on NVIDIA A6000 GPUs. 
After training, we generate 500 radiology reports for each setting.

To evaluate model performance, we adopt our proposed semantics-based evaluation method. 
Specifically, for each generated report, we compute the semantic similarity between every sentence in the candidate report and each sentence in the reference report using GritLM embeddings. 
Sentence pairs exceeding a cosine similarity threshold of 0.8 are considered semantically matched. The proportion of matched sentences is used to assess semantic fidelity. 
In addition, we analyze the distribution of sentence pairs across similarity score ranges to better understand how different training regimes affect the semantic quality and variability of model outputs.




\section{Experimental Results}

\subsection{Readability of Layman-Style Reports}
\begin{table*}[!ht]
\scriptsize
\centering
\begin{tabular}{c|l|c|ccccccccc}
\hline
\multirow{2}{*}{\textbf{Data}} & \multirow{2}{*}{\textbf{Model}} & \multirow{2}{*}{\textbf{\makecell{Easy\\Level}}$\uparrow$} & \multicolumn{9}{c}{\textbf{Level of Grade Required for Reading}$\downarrow$} \\
 &  & & M1 & M2 & M3 & M4 & M5 & M6 & M7 & M8 & M9 \\
\hline
\multirow{4}{*}{\makecell{MIMIC\\CXR}} & (Original) & 43 & 9 & 11 & 11 & 11 & 14 & 5 & 11 & 5 & 11 \\
\cline{2-12}
 & Baseline & 76 & 6 & 8 & 8 & 8 & 9 & 7 & 10 & 5 & \textbf{19} \\
 & LLM1+P1 & 84 & \textbf{5} & 8 & 8 & 7 & \textbf{7} & 7 & \textbf{8} & \textbf{4} & 21 \\
 & LLM1+P2 & \textbf{85} & \textbf{5} & \textbf{7} & \textbf{7} & \textbf{6} & \textbf{7} & \textbf{6} & \textbf{8} & \textbf{4} & \textbf{19} \\
\hline

\end{tabular}
\caption{Readability of Layman-Style Reports. Original represents professional reports. Baseline, LLM1+P1 and LLM1+P2 indicate layman-style reports generated by different LLMs and different prompts.}
\label{tab:read-sixing-1}
\end{table*}
We first evaluated the readability of LLM-generated layman-style radiology reports using two publicly available models, Kimi\footnote{Kimi (\url{www.moonshot.cn})} and DeepSeek\footnote{DeepSeek (\url{www.deepseek.com/})}, on the MIMIC-CXR dataset, denoted as LLM1 and LLM2, respectively. 
To assess readability, we employed a suite of text-statistics-based metrics\footnote{We use the open-source Python library available at \url{pypi.org/project/textstat}}. 
The abbreviations and descriptions of these metrics are listed in Appendix~\ref{app:additional}.
The \texttt{Baseline} approach refers to layman-style reports generated using the prompt provided in Appendix~\ref{prompt:trans} via \texttt{ChatGPT-4o}, while the \texttt{Original} approach corresponds to the professional radiology reports without modification.
In addition to the baseline prompt (\texttt{P1}), we designed an instruction-following prompt (\texttt{P2}) that guides the model to generate layman-style reports based on provided examples. 
An illustration of this prompt is shown in Figure~\ref{fig:prompt-sixing}.
As shown in Table~\ref{tab:read-sixing-1}, the layman-style reports produced by all three LLM approaches demonstrate substantially higher readability than the original professional reports across all evaluation metrics.



\subsection{Limitations of Lexical-based Evaluation}
In this section, we reveal the behavioral differences between lexical-based evaluation metrics and our proposed semantics-based evaluation metric.

To verify the effectiveness of layman-style reports in addressing expression difference and semantic difference issues, we construct two diagnostic subsets: (1) Similar Semantics \& Different Expressions (SS \& DE) and (2) Different Semantics \& Similar Expressions (DS \& SE). The way lexical-based and semantics-based metrics respond to these subsets serves as a characterization of their robustness.

For both raw professional reports and their layman-style counterparts, we compute BLEU, ROUGE, and METEOR scores, along with semantic similarity between candidate and reference sentences, within each diagnostic subset. The results are shown in Table~\ref{tab:score}. In the “DS \& SE” subset, sentence pairs in the professional reports are mistakenly assigned high scores by lexical metrics—for example, 0.644 (BLEU-1), 0.505 (BLEU-2), 0.393 (BLEU-3), and 0.312 (BLEU-4). In contrast, their layman-translated counterparts significantly mitigate this mirage effect, reducing the scores to 0.312, 0.116, 0.064, and 0.042, respectively. Furthermore, our semantics-based metric correctly reflects the lack of semantic similarity in these pairs, with the proportion of sentences scoring above 0.8 dropping to only 2\% and 1\%.

Conversely, in the “SS \& DE” subset, an ideal evaluation metric should be robust to surface-level differences and assign high scores to semantically aligned sentence pairs. However, lexical-based metrics fail to capture this relationship, yielding significantly lower scores for professional report pairs. Our translated layman pairs alleviate this weakness, producing higher perceived scores under lexical metrics. More importantly, the combination of our layman-style dataset and semantics-based metric yields the most robust evaluation: it not only achieves a high proportion of semantically similar pairs (over 50\% scoring above 0.8), but also maintains a small perceptual gap between professional and layman versions.

In summary, lexical-based metrics suffer from inherent limitations, particularly when applied to the highly patterned structure of professional radiology reports. These metrics often fail to reflect the true semantic relationships between sentence pairs—frequently assigning higher scores to DS pairs than to SS pairs. Our layman-style dataset helps correct this imbalance, reversing the trend and enabling lexical metrics to better align with semantic intent. Most importantly, the combination of semantics-based evaluation and layman-style reports provides the most robust and faithful assessment of generated report quality.

\begin{table}[t]
\scriptsize
\centering
\begin{tabular}{cccccccl}
\toprule[1pt]
\multicolumn{1}{c|}{Dataset}
&\multicolumn{2}{c|}{SS\&DE} &\multicolumn{2}{c|}{DS\&SE}
\\
\midrule[1pt]
\multicolumn{1}{c|}{Type} 
&\multicolumn{1}{c|}{raw} &\multicolumn{1}{c|}{layman} &\multicolumn{1}{c|}{raw} &\multicolumn{1}{c|}{layman}
\\
\midrule[1pt]
\multicolumn{1}{c|}{B-1} &\multicolumn{1}{c|}{0.192} &\multicolumn{1}{c|}{0.381} &\multicolumn{1}{c|}{0.644} &\multicolumn{1}{c|}{0.314}
\\
\midrule[1pt]
\multicolumn{1}{c|}{B-2} &\multicolumn{1}{c|}{0.131} &\multicolumn{1}{c|}{0.251} &\multicolumn{1}{c|}{0.505} &\multicolumn{1}{c|}{0.116}
\\
\midrule[1pt]
\multicolumn{1}{c|}{B-3} &\multicolumn{1}{c|}{0.100} &\multicolumn{1}{c|}{0.178} &\multicolumn{1}{c|}{0.393} &\multicolumn{1}{c|}{0.064}
\\
\midrule[1pt]
\multicolumn{1}{c|}{B-4} &\multicolumn{1}{c|}{0.066} &\multicolumn{1}{c|}{0.116} &\multicolumn{1}{c|}{0.312} &\multicolumn{1}{c|}{0.042}
\\
\midrule[1pt]
\multicolumn{1}{c|}{R-1} &\multicolumn{1}{c|}{0.349} &\multicolumn{1}{c|}{0.407} &\multicolumn{1}{c|}{0.622} &\multicolumn{1}{c|}{0.286}
\\
\midrule[1pt]
\multicolumn{1}{c|}{R-2} &\multicolumn{1}{c|}{0.169} &\multicolumn{1}{c|}{0.210} &\multicolumn{1}{c|}{0.399} &\multicolumn{1}{c|}{0.072}
\\
\midrule[1pt]
\multicolumn{1}{c|}{R-L} &\multicolumn{1}{c|}{0.341} &\multicolumn{1}{c|}{0.383} &\multicolumn{1}{c|}{0.581} &\multicolumn{1}{c|}{0.250}
\\
\midrule[1pt]
\multicolumn{1}{c|}{Meteor} &\multicolumn{1}{c|}{0.386} &\multicolumn{1}{c|}{0.452} &\multicolumn{1}{c|}{0.627} &\multicolumn{1}{c|}{0.310}
\\
\midrule[1pt]
\multicolumn{1}{c|}{Semantics} &\multicolumn{1}{c|}{0.5} &\multicolumn{1}{c|}{0.507} &\multicolumn{1}{c|}{0.02} &\multicolumn{1}{c|}{0.01}\\

\hline
\end{tabular}
\caption{BLEU and ROUGE score in professional report and its layman's term. SS\&DE represent similar semantics and different expressions; DS\&SE means different semantics and similar expressions. Semantic scores are calculated with the proportion of semantic similarity over 0.8 among all sentences.}
\label{tab:score}
\end{table}



\subsection{Improving Model Training with Layman’s Terms: Insights from a Scaling Law}
\begin{figure*}[ht]
\centering
\includegraphics[width=0.87\linewidth]{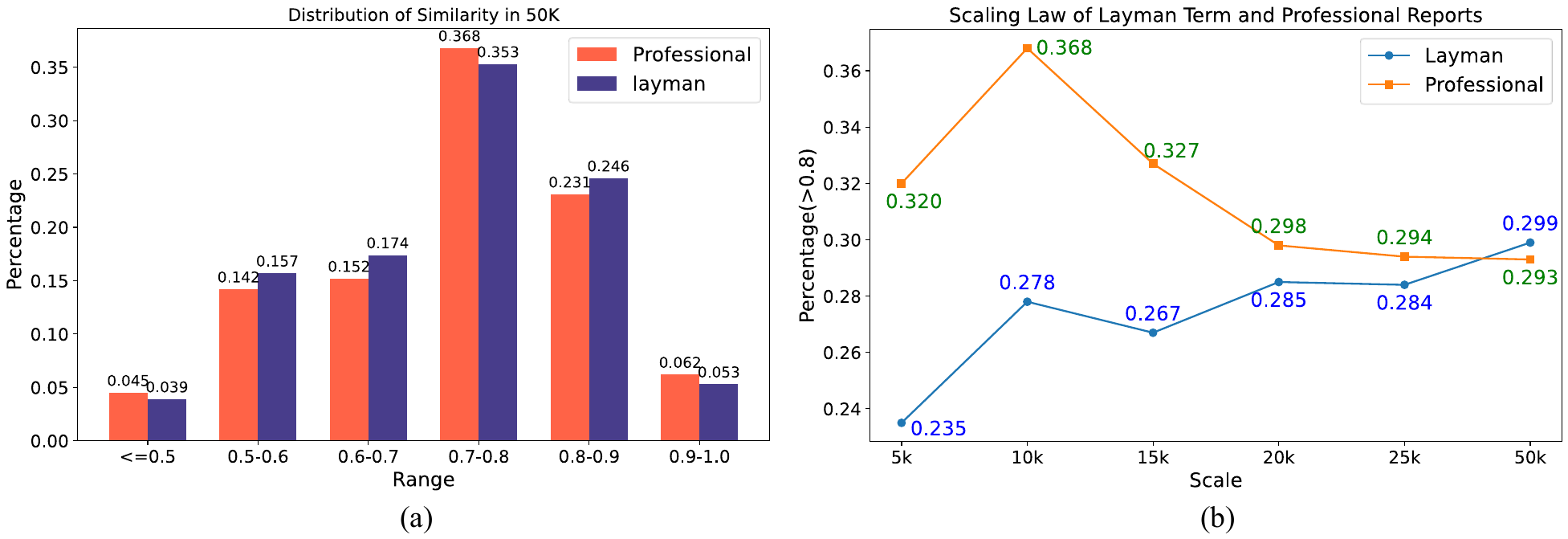}
\caption{Scaling law of the model's semantic understanding by training on report-level datasets.}
\label{fig:label1}
\end{figure*}

To evaluate the impact of training data style on semantic learning, we compare models trained on professional versus layman-style radiology reports using our semantics-based evaluation metric. 
As shown in Figure~\ref{fig:label1}(b), the model trained on layman-style data demonstrates a clear positive scaling law: semantic performance steadily improves as the training set size increases from 5k to 50k. 
In contrast, the model trained on professional reports peaks at 10k samples and declines thereafter, suggesting that prolonged exposure to highly templated language leads to overfitting and reduced semantic generalization. 
Notably, the layman-style dataset starts to outperform professional reports when the training size reaches 50k.

To further assess semantic quality, we analyze the distribution of sentence-level similarity scores under the 50k training setting, as shown in Figure~\ref{fig:label1}(a), with full statistics across all training scales provided in Appendix~\ref{scaling}. 
The layman-style model yields more sentence pairs with high similarity scores (e.g., >0.8), indicating stronger alignment with the reference semantics. 
In contrast, the professional model produces more outputs in the mid-to-low similarity range, reflecting weaker semantic fidelity.

To understand why the model trained on 10k professional reports achieves the highest semantic performance, we conduct further analysis and identify signs of representation collapse. 
Specifically, we compute the pairwise cosine similarity of generated reports on the test set. 
The 10k professional model exhibits an average cosine similarity of 0.893 with a variance of 0.008, suggesting that the model learns to mimic the dominant class (e.g., no findings or normal reports) to minimize loss, rather than capturing diverse semantic content.
In contrast, the 10k layman-style model yields a lower average similarity of 0.802 with a higher variance of 0.012, reflecting greater report diversity and semantic richness.
These findings, combined with the overfitting trend observed in Figure~\ref{fig:label1}, support the conclusion that the layman-style dataset promotes a more robust and natural progression in semantic learning as the dataset scales—unlike the shortcut behavior observed with professional reports. 
Furthermore, we evaluate specialized clinical metrics and find that at the 10k scale, the layman-trained model outperforms its professional counterpart (CheXbert: 0.447 vs. 0.398; RadCliQ-v0: 0.413 vs. 0.405).

\subsection{Evaluating Semantic Fidelity: Human vs. Automated Metrics}
Due to the obscurity of professional radiology reports and the high cost of involving clinicians as annotators, few studies have explored the correlation between human scores and automated metrics such as BLEU in this domain. 
However, it is well documented in other fields that word-overlap-based metrics often fail to capture semantic accuracy and typically exhibit weak correlation with human evaluations. 
Therefore, relying solely on such metrics to assess the quality of generated radiology reports is inadequate.
To enable a fair comparison between models trained on professional and layman-style reports, and to make professional reports more comprehensible to non-clinician human evaluators, we first translate all professional references into layman terms. 
We then recruit three human annotators—fluent English speakers with non-clinical backgrounds—to score the generated reports using a unified evaluation protocol:\textit{``Given the generated text and the reference, calculate the proportion of sentences in the generated text that semantically match each sentence in the reference."} 
This protocol is consistently applied to evaluate both types of model outputs. After collecting scores from all annotators, we compute the final report score by averaging across annotators and across reference-matched sentences. The inter-annotator agreement (IAA), measured by Cohen’s Kappa, is 0.63 for professional reports and 0.58 for layman-style reports, indicating fair to good agreement (0.4–0.75 range). Details about the annotators and scoring procedures are provided in Appendix~\ref{human}.
The correlation results between human evaluations and automated metrics are presented in Table~\ref{tab:cor}. Across the board, reports generated in layman terms show stronger alignment with human judgments. This holds not only for lexical metrics such as BLEU, ROUGE, and METEOR, but also for clinically relevant Clinical Efficacy (CE) metrics, including CheXbert-F1, RadGraph-F1, and RadCliQ. Although CE metrics are designed to assess named entity correctness in medical texts, we find them equally applicable to layman-style reports. Notably, the correlation between CE metrics and human scores is consistently higher for layman-style outputs, reinforcing their semantic fidelity and accessibility.

\begin{table}[t]
\scriptsize
\centering
\begin{tabular}{cccccccl}
\toprule[1pt]

\multicolumn{1}{c|}{Correlation}
&\multicolumn{2}{c|}{Pearson} &\multicolumn{2}{c}{Spearman}
\\
\midrule[1pt]
\multicolumn{1}{c|}{Type}
&\multicolumn{1}{c|}{raw} &\multicolumn{1}{c|}{layman} &\multicolumn{1}{c|}{raw} &\multicolumn{1}{c|}{layman}
\\
\midrule[1pt]
\multicolumn{1}{c|}{B-1} &\multicolumn{1}{c|}{0.533} &\multicolumn{1}{c|}{0.534$\uparrow$} &\multicolumn{1}{c|}{0.536 } &\multicolumn{1}{c|}{0.524} 
\\
\midrule[1pt]
\multicolumn{1}{c|}{B-2} &\multicolumn{1}{c|}{0.526} &\multicolumn{1}{c|}{0.573$\uparrow$} &\multicolumn{1}{c|}{0.532} &\multicolumn{1}{c|}{0.538$\uparrow$} 
\\
\midrule[1pt]
\multicolumn{1}{c|}{B-3} &\multicolumn{1}{c|}{0.480} &\multicolumn{1}{c|}{0.557$\uparrow$} &\multicolumn{1}{c|}{0.502} &\multicolumn{1}{c|}{0.519$\uparrow$} 
\\
\midrule[1pt]
\multicolumn{1}{c|}{B-4} &\multicolumn{1}{c|}{0.420} &\multicolumn{1}{c|}{0.519$\uparrow$} &\multicolumn{1}{c|}{0.450} &\multicolumn{1}{c|}{0.472$\uparrow$} 
\\
\midrule[1pt]
\multicolumn{1}{c|}{R-1} &\multicolumn{1}{c|}{0.543} &\multicolumn{1}{c|}{0.586$\uparrow$} &\multicolumn{1}{c|}{0.550} &\multicolumn{1}{c|}{0.565$\uparrow$} 
\\
\midrule[1pt]
\multicolumn{1}{c|}{R-2} &\multicolumn{1}{c|}{0.430} &\multicolumn{1}{c|}{0.524$\uparrow$} &\multicolumn{1}{c|}{0.441} &\multicolumn{1}{c|}{0.485$\uparrow$} 
\\
\midrule[1pt]
\multicolumn{1}{c|}{R-L} &\multicolumn{1}{c|}{0.526} &\multicolumn{1}{c|}{0.561$\uparrow$} &\multicolumn{1}{c|}{0.532} &\multicolumn{1}{c|}{0.534$\uparrow$} 
\\
\midrule[1pt]
\multicolumn{1}{c|}{Meteor} &\multicolumn{1}{c|}{0.527} &\multicolumn{1}{c|}{0.586$\uparrow$} &\multicolumn{1}{c|}{0.538} &\multicolumn{1}{c|}{0.556$\uparrow$} 
\\
\midrule[1pt]
\multicolumn{1}{c|}{Semantics} &\multicolumn{1}{c|}{0.559} &\multicolumn{1}{c|}{0.601$\uparrow$} &\multicolumn{1}{c|}{0.558} &\multicolumn{1}{c|}{0.576$\uparrow$} 
\\
\midrule[1pt]
\multicolumn{1}{c|}{Chexbert} &\multicolumn{1}{c|}{0.570} &\multicolumn{1}{c|}{0.600$\uparrow$} &\multicolumn{1}{c|}{0.620} &\multicolumn{1}{c|}{0.703$\uparrow$} 
\\
\midrule[1pt]
\multicolumn{1}{c|}{Radgraph} &\multicolumn{1}{c|}{0.521} &\multicolumn{1}{c|}{0.652$\uparrow$} &\multicolumn{1}{c|}{0.536} &\multicolumn{1}{c|}{0.658$\uparrow$} 
\\
\midrule[1pt]
\multicolumn{1}{c|}{RadCliQ-v0} &\multicolumn{1}{c|}{0.616} &\multicolumn{1}{c|}{0.710$\uparrow$} &\multicolumn{1}{c|}{0.633} &\multicolumn{1}{c|}{0.724$\uparrow$} 
\\
\midrule[1pt]
\multicolumn{1}{c|}{RadCliQ-v1} &\multicolumn{1}{c|}{0.613} &\multicolumn{1}{c|}{0.719$\uparrow$} &\multicolumn{1}{c|}{0.630} &\multicolumn{1}{c|}{0.728$\uparrow$} 
\\
\bottomrule[1pt]
\end{tabular}
\caption{The correlation of automated metrics (BLEU, ROUGE and semantic scores) and human evaluators, for both professional reports and their layman's terms counterpart. Semantic scores are calculated with the proportion of semantic similarity over 0.8 among all sentences.}
\label{tab:cor}
\end{table}

\subsection{Case Study}
Table~\ref{case study} presents several sentence-level examples demonstrating how translating professional radiology terminology into layman’s language can substantially improve clarity and patient understanding.
For instance, the clinical term \textit{pleural effusion} is rephrased as \textit{extra fluid around the lungs}, offering a more intuitive explanation.
Similarly, \textit{bibasilar atelectasis}, which may be obscure or confusing to non-experts, becomes \textit{collapsed lung areas}, conveying the concept in simpler terms.
These examples highlight the value of plain language in enhancing communication and promoting patient comprehension in medical settings.

\begin{table}[ht]
\small
\centering
\begin{tabular}{cccccccl}
\toprule[1pt]
\multicolumn{1}{c|}{original}
&\multicolumn{1}{c|}{layman}
\\
\midrule[1pt]
\multicolumn{1}{c|}{\begin{tabular}{p{3cm}} 
\sethlcolor{blue!25}
Both lung fields are \hl{clear}\end{tabular}} &\multicolumn{0}{c|}{\begin{tabular}{p{3cm}} 
\sethlcolor{blue!25}
Both lungs look \hl{healthy with no problems}\end{tabular}} 
\\
\midrule[1pt]
\multicolumn{1}{c|}{\begin{tabular}{p{3cm}} 
\sethlcolor{blue!25}
No evidence of \hl{pleural effusion}\end{tabular}} &\multicolumn{0}{c|}{\begin{tabular}{p{3cm}} 
\sethlcolor{blue!25}
There is no \hl{extra fluid around the lungs}\end{tabular}} 
\\
\midrule[1pt]
\multicolumn{1}{c|}{\begin{tabular}{p{3cm}} 
\sethlcolor{blue!25}
The chest x-ray shows \hl{subtle patchy} lateral left lower lobe opacities, which are most likely vascular structures and deemed stable with no definite new \hl{focal consolidation}\end{tabular}} &\multicolumn{0}{c|}{\begin{tabular}{p{3cm}} 
\sethlcolor{blue!25}
The x-ray shows \hl{faint cloudy spots} in the lower part of the left lung, likely blood vessels, and overall stable with no new clear \hl{lung infection}\end{tabular}} 
\\
\midrule[1pt]
\multicolumn{1}{c|}{\begin{tabular}{p{3cm}} 
\sethlcolor{blue!25}
Overall impression suggests appropriate positioning of the tubes and \hl{bibasilar atelectasis}, along with findings consistent with small bowel obstruction\end{tabular}} &\multicolumn{0}{c|}{\begin{tabular}{p{3cm}} 
\sethlcolor{blue!25}
The overall impression suggests proper placement of tubes and some \hl{collapsed lung areas}, along with signs of small bowel obstruction\end{tabular}} 
\\
\midrule[1pt]
\multicolumn{1}{c|}{\begin{tabular}{p{3cm}} 
\sethlcolor{blue!25}
However, \hl{cephalization of engorged pulmonary vessels} has probably improved\end{tabular}} &\multicolumn{0}{c|}{\begin{tabular}{p{3cm}} 
\sethlcolor{blue!25}
\hl{The congested blood vessels} in the lungs have likely improved\end{tabular}} 
\\

\bottomrule[1pt]
\end{tabular}
\caption{Examples from the sentence-level dataset.}
\label{case study}
\end{table}

\section{Conclusion}
In this paper, we presented the Layman’s RRG framework to jointly address the challenges of accessibility and robustness in radiology report generation. At the core of our framework are two high-quality layman-style datasets—at the sentence and report levels—constructed through a rigorous generation and self-refinement pipeline. These datasets serve as the foundation for both evaluation and training. Building on this, we introduced a semantics-based evaluation method that, when paired with our sentence-level dataset, mitigates the overestimated scores produced by traditional word-overlap metrics and more accurately captures the semantic quality of generated reports. Furthermore, we proposed a layman-guided training strategy utilizing the report-level dataset, which enhances the model’s semantic understanding and exhibits a positive scaling behavior, where performance continues to improve as the training data grows.
Collectively, these contributions provide a foundation for building radiology report generation systems that are not only semantically faithful, but also more accessible to patients and non-experts.


\section*{Ethics Statement}

In this paper, we introduce a Layman RRG framework for radiology report generation and evaluation. The advantage of our framework is that it is better for models to enhance the understanding on the semantics, as well as provide a more robust evaluation framework. However, a potential downside is that some layman's terms may express inappropriate or offensive meanings because of the hallucination issues of LLMs. Therefore, it is crucial to carefully review the content of training datasets prior to training the layman models to mitigate this issue.

\section*{Limitations}

Although our Layman RRG framework could provide a promising training process and provide a robust evaluation process, it has certain limitations. Primarily, as we utilized \texttt{GPT-4o} to translate the professional reports to layman's terms and proceed a strict modification process to improve the quality of translated layman's term, it may also include a few of professional reports that do not translate perfectly. In future work, we will focus more on continuing to improve the quality of translated reports.


\bibliography{custom}

\appendix

\section{Appendix}
\label{sec:appendix}

\subsection{Prompt for Translation}
\label{prompt:trans}

\textit{
Given a series of sentences that are split from radiology reports. \\\\
Sentences:\\
\{placeholder for 50 sentences\}\\\\
Please finish the following tasks. \\
Tasks:\\
1. Translation: Please translate each sentence into plain language that is easy to understand. You must translate all the sentences. \\\\
For each task, return a dict. Here are some examples:\\
Task 1:\\
```json\\
\{\\
"0": "No signs of infection, fluid, or air outside of the lung—everything looks normal.", \\
"1": "The unclear spots seen in both lungs are most likely just shadows from nipples.",\\
...\\
\}\\
```\\
}

\subsection{Prompt for Refinement}
\label{prompt:refine}

\textit{
Given a series of Original sentences that are split from radiology reports and their translated layman's terms sentence.\\\\
Original Sentences:\\
\{placeholder for 50 sentences\}\\\\
Translated Layman's Term:\\
\{placeholder for 50 sentences\}\\\\
Please finish the following tasks. \\
Tasks:\\
1. Check and Modification: Please check if the translated sentence is semantically consistent and has the same detailed description as the given original sentence. If it is, make no changes; otherwise, make modifications. \\\\
For each task, return a dict. Here are some examples:\\
Task 1:\\
```json\\
\{\\
"0": "No signs of infection, fluid, or air outside of the lung—everything looks normal.", \\
"1": "The unclear spots seen in both lungs are most likely just shadows from nipples.",\\
...\\
\}\\
```\\
}
\\

\subsection{Dataset}

In this part, we outline the statistics of our datasets as follows in the Table ~\ref{dataset:sta}.

\begin{table}[htb]
\scriptsize \centering
\resizebox{0.99\linewidth}{!}{
\begin{tabular}{l|cc}
\toprule
\textbf{Datasets}&\textbf{Sentence-level}&\textbf{Report-Level}\\
\midrule
\textbf{\# Numbers} & 50000  &50000 \\
\midrule
\textbf{Avg. \# Words per sample} & 28.68 & 101.45  \\
\textbf{Avg. \# Sentences per sample} & 1 & 5.05  \\

\bottomrule
\end{tabular}
}
\caption{Data statistics of the sentence-level and report-level dataset.}
\label{dataset:sta}
\end{table}

\subsection{Dataset Generation and Refinement Algorithm}
\label{alg:refine}
The Dataset Generation and Refinement Algorithm is shown as Algorithm~\ref{alg:dataset}.

\begin{algorithm}[t]
\scriptsize
\caption{Dataset Generation and Refinement}
\label{alg:dataset}
\begin{algorithmic}[1]
\REQUIRE A set of $n$ data items $D = \{d_1, d_2, \ldots, d_n\}$, a threshold $\theta$ for semantic similarity
\ENSURE Translated set $T = \{t_1, t_2, \ldots, t_n\}$ where each $t_i$ is a valid translation of $d_i$
\FOR{$i = 1$ to $n$}
    \REPEAT
        \STATE $t_i \leftarrow \text{LLM-Translate}(d_i)$
        \STATE $sim \leftarrow \text{Semantic-Similarity}(d_i, t_i)$
        \STATE $correct \leftarrow \text{LLM-Check-Translation}(d_i, t_i)$
    \UNTIL{$sim \geq \theta$ \AND $correct$}
\ENDFOR
\STATE \textbf{return} $T$
\end{algorithmic}
\end{algorithm}

\subsection{Candidate Report Evaluation using GRITLM and Layman Term Replacement}
\label{app:eva}

The Candidate Report Evaluation using GRITLM and Layman Term Replacement is shown as Algorithm~\ref{alg:eva}.
\begin{algorithm}[ht]
\small
\caption{Candidate Report Evaluation using GRITLM and Layman Term Replacement}
\label{alg:eva}
\begin{algorithmic}[1]
\REQUIRE Candidate report $C$, Reference report $R$, Sentence-level dataset $S$, Semantic similarity threshold $\theta = 0.8$
\ENSURE Proportion of sentences in $C$ and $R$ with semantic similarity $\geq \theta$ after replacement, BLEU, ROUGE, and Meteor scores
\STATE $C_s \leftarrow \text{Split-Sentences}(C)$
\STATE $R_s \leftarrow \text{Split-Sentences}(R)$
\FOR{each sentence $c_i \in C_s$}
    \STATE $max\_sim \leftarrow 0$
    \FOR{each sentence $s_j \in S$}
        \STATE $sim \leftarrow \text{GRITLM-Similarity}(c_i, s_j)$
        \IF{$sim > max\_sim$}
            \STATE $max\_sim \leftarrow $ sim
            \STATE $replacement \leftarrow \text{Layman-Term}(s_j)$
        \ENDIF
    \ENDFOR
    \STATE $c_i \leftarrow replacement$
\ENDFOR

\FOR{each sentence $r_i \in R_s$}
    \STATE $max\_sim \leftarrow 0$
    \FOR{each sentence $s_j \in S$}
        \STATE $sim \leftarrow \text{GRITLM-Similarity}(r_i, s_j)$
        \IF{$sim > max\_sim$}
            \STATE $max\_sim \leftarrow $ sim
            \STATE $replacement \leftarrow \text{Layman-Term}(s_j)$
        \ENDIF
    \ENDFOR
    \STATE $r_i \leftarrow replacement$
\ENDFOR

\STATE $similar\_count \leftarrow 0$
\FOR{each sentence $c_i \in C_s$}
    \FOR{each sentence $r_i \in R_s$}
        \STATE $sim \leftarrow \text{GRITLM-Similarity}(c_i, r_i)$
        \IF{$sim \geq \theta$}
            \STATE $similar\_count \leftarrow similar\_count + 1$
            \STATE \textbf{break}
        \ENDIF
    \ENDFOR
\ENDFOR

\STATE $proportion \leftarrow \frac{similar\_count}{|C_s|}$

\STATE $BLEU \leftarrow \text{Compute-BLEU}(C_s, R_s)$
\STATE $ROUGE \leftarrow \text{Compute-ROUGE}(C_s, R_s)$
\STATE $Meteor \leftarrow \text{Compute-Meteor}(C_s, R_s)$
\STATE \textbf{return} $proportion, BLEU, ROUGE, Meteor$
\end{algorithmic}
\end{algorithm}

\subsection{Refinement Rate}
\label{appendix: refinement}

\begin{figure}[H]
\centering
\small
\includegraphics[width=1\linewidth]{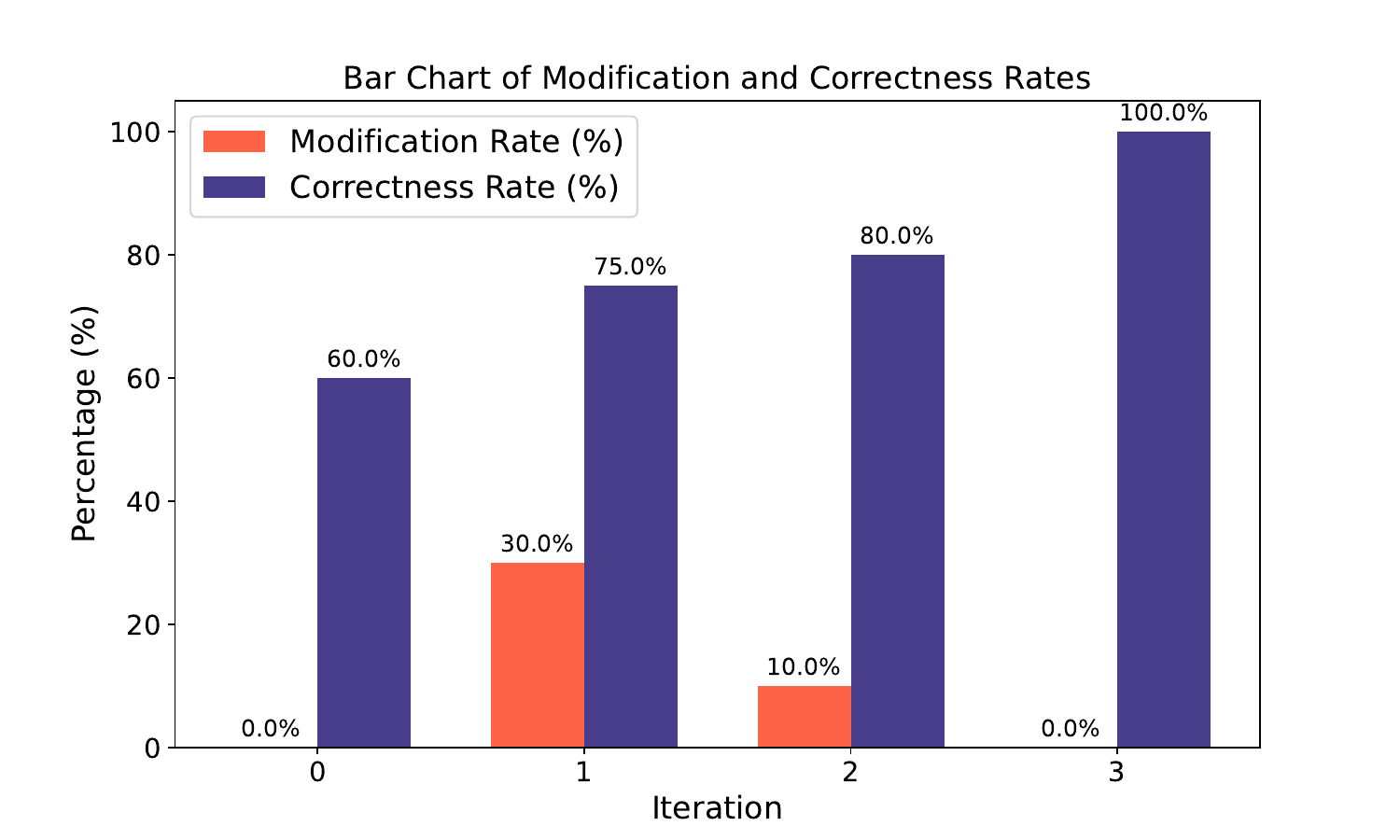}
\caption{Refinement}
\label{refine}
\end{figure}

In this section, we examine a subset of 100 samples to analyze the refinement process, observing both the accuracy proportion at each stage and the sentence modification rate per step. As illustrated in Figure~\ref{refine}, the refinement process concludes after three iterations.

\subsection{Analysis of Refinement Step}

As mentioned in the early parts, our data generation pipeline leverages a rigorous refinement process. This includes a LLM self-refinement module and an embedding model to assess semantic similarity.

Here, we present an example going through 4 steps in the refinement process. As detailed in Table ~\ref{tab:refine}, the example includes the translated report at each step and the calculation of semantic similarity between each sentence in the original professional report and the corresponding sentence in layman's terms. Step 0 is the raw professional report that requires translation, and Steps 1-3 present the reports translated to layman's terms. The red numbers display the semantic similarity. It is evident that the semantic similarity increases in each step and remains unchanged at the third step, signifying the conclusion of the refinement process. This analysis demonstrates that the refinement process effectively enhances the quality of the translated layman's reports.

\begin{table}[ht]
\scriptsize
\centering
\begin{tabular}{cccccccl}
\toprule[1pt]
\multicolumn{1}{c|}{Step}
&\multicolumn{1}{c|}{Report}
\\
\midrule[1pt]
\multicolumn{1}{c|}{0} &\multicolumn{0}{c|}{\begin{tabular}{p{5.8cm}} Subtle rounded nodular opacity projecting over both lung bases which could represent nipple shadows. Recommend repeat with nipple markers to confirm and exclude underlying pulmonary nodule. Subtle bibasilar opacities likely represent atelectasis or aspiration. No evidence of pneumonia.\end{tabular}} 
\\
\midrule[1pt]
\multicolumn{1}{c|}{1} &\multicolumn{0}{c|}{\begin{tabular}{p{5.8cm}} There are some unclear spots in the lower parts of both lungs which might just be shadows caused by nipples \textcolor{red}{(0.776)}. We recommend doing another x-ray using nipple markers to be sure \textcolor{red}{(0.731)}. There are also subtle changes in the lower lungs likely due to collapsed lung areas or inhaled food/liquid \textcolor{red}{(0.704)}. No signs of pneumonia \textcolor{red}{(0.971)}.
\end{tabular}} 
\\
\midrule[1pt]
\multicolumn{1}{c|}{2} &\multicolumn{0}{c|}{\begin{tabular}{p{5.8cm}} The unclear spots seen in both lung bases are most likely just shadows from nipples \textcolor{red}{(0.778$\uparrow$)}. We recommend a repeat x-ray with nipple markers to confirm and exclude any underlying lung nodules \textcolor{red}{(0.911$\uparrow$)}. There are also subtle changes in the lower lungs likely due to collapsed lung areas or inhalation of food/liquid \textcolor{red}{(0.712$\uparrow$)}. No evidence of pneumonia \textcolor{red}{(0.999$\uparrow$)}.
\end{tabular}} 
\\
\midrule[1pt]
\multicolumn{1}{c|}{3} &\multicolumn{0}{c|}{\begin{tabular}{p{5.8cm}} The unclear spots seen in both lung bases are most likely just shadows from nipples. We recommend a repeat x-ray with nipple markers to confirm and exclude any underlying lung nodules. There are also subtle changes in the lower lungs likely due to collapsed lung areas or inhalation of food/liquid. No evidence of pneumonia. \textcolor{red}{(Refinement ends)}
\end{tabular}} 
\\
\bottomrule[1pt]
\end{tabular}
\caption{The expression of an example going through the refinement process. }
\label{tab:refine}
\end{table}

\subsection{Instruction Tuning}
\begin{table}[ht]
\centering
\begin{tabular}{l|c}
\toprule[1pt]
\multicolumn{1}{c|}{Training set}
&\multicolumn{1}{c|}{Similarity >0.8}
\\
\midrule[1pt]
\multicolumn{1}{c|}{professional 50k} &\multicolumn{0}{c|}{0.293} 
\\
\midrule[1pt]
\multicolumn{1}{c|}{layman 50k} &\multicolumn{0}{c|}{0.299} 
\\
\midrule[1pt]
\multicolumn{1}{c|}{professional + layman 100k} &\multicolumn{0}{c|}{0.323} 
\\
\bottomrule[1pt]
\end{tabular}
\caption{Instruction Tuning}
\label{tab:instruct}
\end{table}
We further ran an initial experiment for the new application, by concatenating the 50k professional dataset and the 50k layman’s dataset, yielding a 100k two-class instruction tuning training set. We hypothesize that seeing both versions with different wordings would encourage the model to pick up the semantic overlaps between the two datasets.

For the two datasets, we prepend their corresponding instruction to the example: “Given this X-ray image, generate a professional radiology report.”, “Given this X-ray image, generate a radiology report in layman’s terms.” and in inference, we prepend the same instructions based on our need. The experiments took 5 days on 4 A6000 GPUs.

In Table~\ref{tab:instruct}, we reported the model performance on three settings: 1) trained professional \& inference professional 2) trained layman \& inference layman 3) trained both \& inference professional. We show the percentage of generated reports that have over 0.8 cosine similarity with the groundtruth reports for each setting, aligning with the setting in Figure 3 (right) in the paper.

As shown in the results, the instruction-tuned model, when exposed to both professional and layman reports in the training, can generate a higher percentage of professional reports that are more semantically aligned with the groundtruth. This has indicated that the model is able to pick up semantic hints from the layman’s dataset in the training to enhance its professional report generation. More importantly, this new unified model can generate both professional and layman’s reports when provided with the instructions.




\subsection{Case Study}
\label{case}

In this section, we provide more examples from sentence-level dataset and report-level dataset. The Table~\ref{case1} include some examples in the sentence-level dataset and Table~\ref{case2} present samples selected from the report-level dataset.

\begin{table}[ht]
\scriptsize
\centering
\begin{tabular}{cccccccl}
\toprule[1pt]
\multicolumn{1}{c|}{raw}
&\multicolumn{1}{c|}{layman}
\\
\midrule[1pt]
\multicolumn{1}{c|}{\begin{tabular}{p{3cm}} Both lung fields are clear\end{tabular}} &\multicolumn{0}{c|}{\begin{tabular}{p{3cm}} Both lungs look healthy with no problems\end{tabular}} 
\\
\midrule[1pt]
\multicolumn{1}{c|}{\begin{tabular}{p{3cm}} No evidence of pleural effusion\end{tabular}} &\multicolumn{0}{c|}{\begin{tabular}{p{3cm}} There is no extra fluid around the lungs\end{tabular}} 
\\
\midrule[1pt]
\multicolumn{1}{c|}{\begin{tabular}{p{3cm}} The chest x-ray shows subtle patchy lateral left lower lobe opacities, which are most likely vascular structures and deemed stable with no definite new focal consolidation\end{tabular}} &\multicolumn{0}{c|}{\begin{tabular}{p{3cm}} The x-ray shows faint cloudy spots in the lower part of the left lung, likely blood vessels, and overall stable with no new clear lung infection\end{tabular}} 
\\
\midrule[1pt]
\multicolumn{1}{c|}{\begin{tabular}{p{3cm}} The impression states that the opacities are bilateral and indicative of an infection that requires follow up attention to ensure resolution\end{tabular}} &\multicolumn{0}{c|}{\begin{tabular}{p{3cm}} The impression notes the cloudy spots are in both lungs, likely indicating an infection that needs follow-up to ensure it's resolved\end{tabular}} 
\\
\midrule[1pt]
\multicolumn{1}{c|}{\begin{tabular}{p{3cm}} Overall impression suggests appropriate positioning of the tubes and bibasilar atelectasis, along with findings consistent with small bowel obstruction\end{tabular}} &\multicolumn{0}{c|}{\begin{tabular}{p{3cm}} The overall impression suggests proper placement of tubes and some collapsed lung areas, along with signs of small bowel obstruction\end{tabular}} 
\\
\midrule[1pt]
\multicolumn{1}{c|}{\begin{tabular}{p{3cm}} A mildly displaced fracture of the right anterior sixth rib and possible additional right anterior seventh rib fracture are noted\end{tabular}} &\multicolumn{0}{c|}{\begin{tabular}{p{3cm}} There is a slightly displaced fracture of the right front sixth rib and possibly another right front seventh rib fracture\end{tabular}} 
\\
\midrule[1pt]
\multicolumn{1}{c|}{\begin{tabular}{p{3cm}} There is increased soft tissue density at the left hilum and a fiducial seed is seen in an unchanged position\end{tabular}} &\multicolumn{0}{c|}{\begin{tabular}{p{3cm}} Increased tissue density is seen at the left lung root and a tracking marker is in the same place as before\end{tabular}} 
\\
\midrule[1pt]
\multicolumn{1}{c|}{\begin{tabular}{p{3cm}} However, cephalization of engorged pulmonary vessels has probably improved\end{tabular}} &\multicolumn{0}{c|}{\begin{tabular}{p{3cm}} The congested blood vessels in the lungs have likely improved\end{tabular}} 
\\
\midrule[1pt]
\multicolumn{1}{c|}{\begin{tabular}{p{3cm}} Moderate bilateral layering pleural effusions are also present along with a notable compression deformity of a lower thoracic vertebral body, without information about the age of the patient\end{tabular}} &\multicolumn{0}{c|}{\begin{tabular}{p{3cm}} Moderate fluid in both pleura is seen along with a compression deformity in a lower chest spine bone, without age information on the patient\end{tabular}} 
\\
\midrule[1pt]
\multicolumn{1}{c|}{\begin{tabular}{p{3cm}} The chest x-ray image reveals worsening diffuse alveolar consolidations with air bronchograms, particularly in the right apex and entire left lung\end{tabular}} &\multicolumn{0}{c|}{\begin{tabular}{p{3cm}} The x-ray shows worsening of diffuse lung cloudiness with air-filled bronchial tubes, especially in the right lung apex and the entire left lung\end{tabular}} 
\\

\bottomrule[1pt]
\end{tabular}
\caption{Some examples of sentence-level dataset.}
\label{case1}
\end{table}

\begin{table}[ht]
\scriptsize
\centering
\begin{tabular}{cccccccl}
\toprule[1pt]
\multicolumn{1}{c|}{raw}
&\multicolumn{1}{c|}{layman}
\\
\midrule[1pt]
\multicolumn{1}{c|}{\begin{tabular}{p{3cm}} Bilateral nodular opacities, which most likely represent nipple shadows, are observed. There is no focal consolidation, pleural effusion, or pneumothorax. Cardiomediastinal silhouette is normal, and there is no acute cardiopulmonary process. Clips project over the left lung, potentially within the breast, and the imaged upper abdomen is unremarkable. Chronic deformity of the posterior left sixth and seventh ribs is noted.\end{tabular}} &\multicolumn{0}{c|}{\begin{tabular}{p{3cm}} There are spots seen in both lungs that are likely just nipple shadows. There is no evidence of a specific infection, fluid in the lungs, or air outside the lungs. The shape of the heart and area around it looks normal. There are no immediate heart or lung issues. There are surgical clips in the area of the left lung, likely in the breast, and the upper abdomen appears normal. There is a long-term deformity of the sixth and seventh ribs on the left side.\end{tabular}} 
\\
\midrule[1pt]
\multicolumn{1}{c|}{\begin{tabular}{p{3cm}} The chest x-ray shows normal cardiac, mediastinal, and hilar contours with clear lungs and normal pulmonary vasculature. No pleural effusion or pneumothorax is present. However, multiple clips are seen projecting over the left breast, and remote left-sided rib fractures are also demonstrated. The impression is that there is no acute cardiopulmonary abnormality detected.\end{tabular}} &\multicolumn{0}{c|}{\begin{tabular}{p{3cm}} The chest x-ray shows a normal heart shape and clear lungs with no fluid or air outside the lungs. There are multiple surgical clips seen in the left breast area, and old rib fractures on the left side. There are no immediate heart or lung problems detected.\end{tabular}} 
\\
\midrule[1pt]
\multicolumn{1}{c|}{\begin{tabular}{p{3cm}} The chest x-ray shows no evidence of focal consolidation, effusion, or pneumothorax, and the cardiomediastinal silhouette is normal. Multiple clips projecting over the left breast and remote left-sided rib fractures are noted. No free air below the right hemidiaphragm is seen. The impression is that there is no acute intrathoracic process.\end{tabular}} &\multicolumn{0}{c|}{\begin{tabular}{p{3cm}} The chest x-ray does not show any specific lung infection, fluid, or air outside the lungs. The heart and surrounding area appear normal. Multiple surgical clips are seen in the left breast area, and old rib fractures on the left side are noted. There is no free air under the right side of the diaphragm. There are no immediate issues inside the chest.\end{tabular}} 
\\

\bottomrule[1pt]
\end{tabular}
\caption{Some examples of report-level dataset.}
\label{case2}
\end{table}

\subsection{Additional Experiments}
\label{app:additional}

\begin{figure*}[ht!]
\scriptsize
\centering 
\includegraphics[width=1\linewidth]{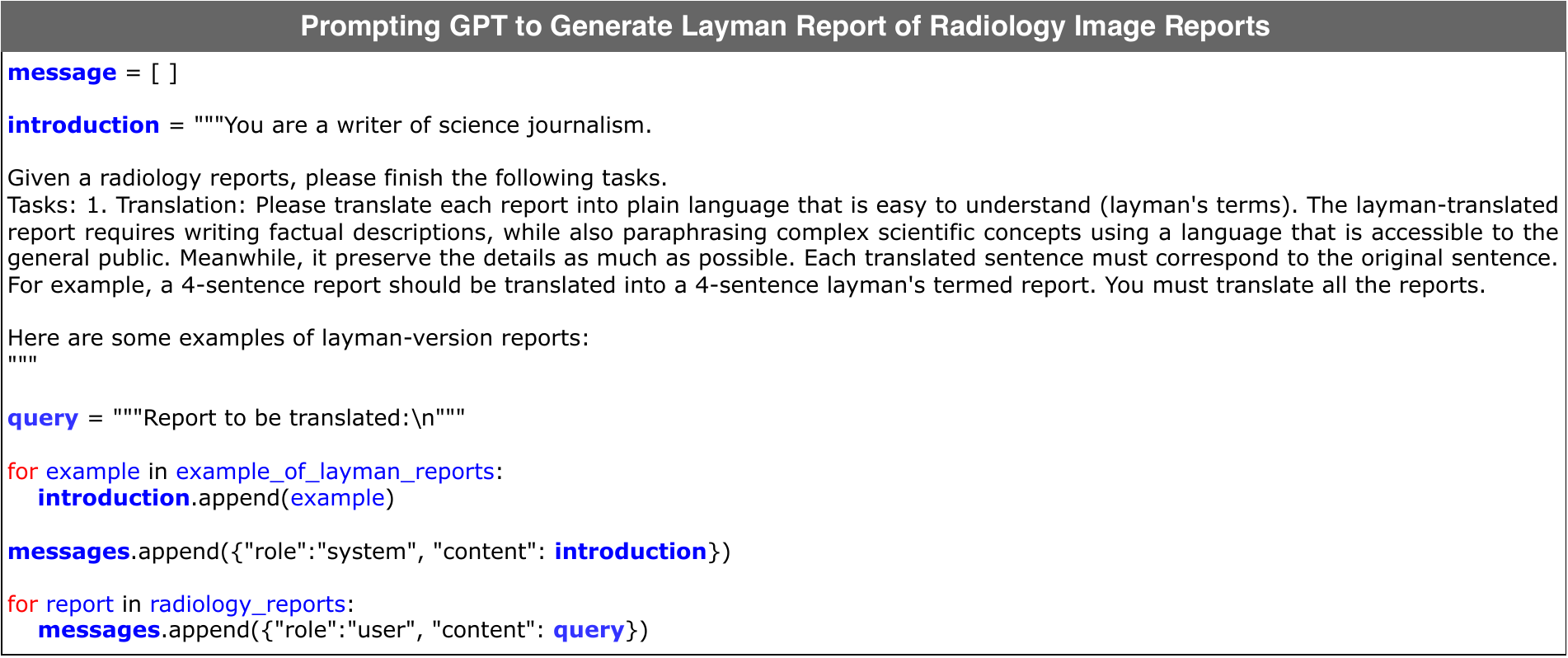}
\caption{Example of prompting GPT to generate the layman report of the radiology image reports.}
\label{fig:prompt-sixing}
\end{figure*}

We also tested the LLM-based approach using two different open-access ChatGPTs\footnote{Kimi (\url{www.moonshot.cn}) and DeepSeek (\url{www.deepseek.com/})} in both MIMIC CXR and PadChest (English translated) datasets, denoted as LLM1 and LLM2, respectively.
\texttt{Baseline} approach in MIMIC CXR dataset indicates the layman reports which using prompts provided in \ref{prompt:trans}.
\texttt{(Original)} approach in MIMIC CXR and PadChest indicate the original radiology reports. We also reported their readability scores. 
Apart from the baseline prompt (denoted as \texttt{P1}), a instruction-following prompt (denoted as \texttt{P2}) is designed for GPT to generate layman report by examples provided. An example is shown in Fig.~\ref{fig:prompt-sixing}.

The evaluation metrics are in three types: i) Clinical accuracy, ii) Relevance, and iii) Readability. 
For Readability, a set of text statistics metrics\footnote{The open-source Python library is provided on \url{pypi.org/project/textstat}} to be used. Their abbreviation and the corresponding metrics are listed below:
\begin{itemize}
    \item \texttt{Easy}: The Flesch Reading Ease formula
    \item \texttt{M1}: The Flesch-Kincaid Grade Level
    \item \texttt{M2}: The Fog Scale (Gunning FOG Formula)
    \item \texttt{M3}: The SMOG Index
    \item \texttt{M4}: Automated Readability Index
    \item \texttt{M5}: The Coleman-Liau Index
    \item \texttt{M6}: Linsear Write Formula
    \item \texttt{M7}: Dale-Chall Readability Score
    \item \texttt{M8}: Spache Readability Formula
    \item \texttt{M9}: McAlpine EFLAW Readability Score
\end{itemize}
The experimental results are provided in Table~\ref{tab:acc-sixing} and Table~\ref{tab:read-sixing}.

\begin{table*}[!ht]
\scriptsize
\centering
\begin{tabular}{c|l|cccccc|cccc}
\hline
\multirow{3}{*}{\textbf{Data}} & \multicolumn{1}{l|}{\multirow{3}{*}{\textbf{Model}}} & \multicolumn{6}{c|}{\textbf{Clinical Accuracy}} & \multicolumn{4}{c}{\textbf{Relevance}} \\
 & & \multicolumn{3}{c}{Chexbert-F1} & \multicolumn{3}{c|}{RadGraph-F1} & \multirow{2}{*}{B.} & \multirow{2}{*}{M.} & \multirow{2}{*}{R.} & \multirow{2}{*}{Sem.} \\
 & & Acc & Micro & Macro & R1 & R2 & R3 &  &  &  & \\
\hline
\multirow{3}{*}{\makecell{MIMIC\\CXR}} & Baseline & 0.737 & 0.576 & 0.076 & 0.026 & 0.023 & 0.016 & 0.073 & 0.299 & 0.337 & 0.577 \\
& LLM1+P1 & 0.771 & 0.602 & 0.086 & 0.012 & 0.010 & 0.007 & 0.085 & 0.366 & \textbf{0.348} & 0.587 \\
 & LLM1+P2 & \textbf{0.846} & \textbf{0.776} & \textbf{0.138} & \textbf{0.028} & \textbf{0.024} & \textbf{0.017} & \textbf{0.087} & \textbf{0.384} & 0.347 & \textbf{0.758} \\
\hline
\hline
\multirow{4}{*}{PadChest}
& LLM1+P1 & 0.918 & 0.655 & 0.060 & 0.058 & 0.039 & 0.030 & \textbf{0.068} & \textbf{0.436} & \textbf{0.251} & 0.685 \\
& LLM1+P2 & \textbf{0.940} & \textbf{0.748} & \textbf{0.075} & \textbf{0.065} & \textbf{0.041} & \textbf{0.029} & 0.065 & 0.421 & 0.244 & \textbf{0.778} \\
\cline{2-12}
& LLM2+P1 & \textbf{0.945} & \textbf{0.746} & \textbf{0.074} & 0.095 & 0.073 & 0.061 & 0.084 & 0.389 & 0.267 & 0.778 \\
& LLM2+P2 & 0.937 & 0.736 & 0.073 & \textbf{0.153} & \textbf{0.134} & \textbf{0.122} & \textbf{0.188} & \textbf{0.497} & \textbf{0.373} & \textbf{0.792} \\
\hline
\end{tabular}
\caption{Clinical Accuracy and Relevance of Layman-style reports on MIMIC-CXR and PadChest Dataset. Baseline, LLM1+P1 and LLM1+P2 indicate layman-style reports generated by different LLMs and different prompts.}
\label{tab:acc-sixing}
\end{table*}

\begin{table*}[!ht]
\scriptsize
\centering
\begin{tabular}{c|l|c|ccccccccc}
\hline
\multirow{2}{*}{\textbf{Data}} & \multirow{2}{*}{\textbf{Model}} & \multirow{2}{*}{\textbf{\makecell{Easy\\Level}}$\uparrow$} & \multicolumn{9}{c}{\textbf{Level of Grade Required for Reading}$\downarrow$} \\
 &  & & M1 & M2 & M3 & M4 & M5 & M6 & M7 & M8 & M9 \\
\hline
\multirow{4}{*}{\makecell{MIMIC\\CXR}} & (Original) & 43 & 9 & 11 & 11 & 11 & 14 & 5 & 11 & 5 & 11 \\
\cline{2-12}
 & Baseline & 76 & 6 & 8 & 8 & 8 & 9 & 7 & 10 & 5 & \textbf{19} \\
 & LLM1+P1 & 84 & \textbf{5} & 8 & 8 & 7 & \textbf{7} & 7 & \textbf{8} & \textbf{4} & 21 \\
 & LLM1+P2 & \textbf{85} & \textbf{5} & \textbf{7} & \textbf{7} & \textbf{6} & \textbf{7} & \textbf{6} & \textbf{8} & \textbf{4} & \textbf{19} \\
\hline \hline
\multirow{5}{*}{PadChest} & (Original) & 26 & 12 & 14 & 4 & 14 & 16 & 5 & 14 & 6 & 10 \\
\cline{2-12}
 & LLM1+P1 & 69 & 7 & 9 & 4 & \textbf{8} & 9 & \textbf{7} & \textbf{9} & 5 & 19 \\
 & LLM1+P2 & \textbf{73} & \textbf{6} & \textbf{8} & \textbf{3} & \textbf{8} & \textbf{8} & \textbf{7} & \textbf{9} & \textbf{4} & \textbf{18} \\
 \cline{2-12}
 & LLM2+P1 & \textbf{68} & \textbf{8} & \textbf{9} & 4 & \textbf{9} & \textbf{10} & 8 & \textbf{10} & \textbf{5} & 21 \\
 & LLM2+P2 & 64 & \textbf{8} & 10 & \textbf{3} & \textbf{9} & \textbf{10} & \textbf{7} & 11 & \textbf{5} & \textbf{18} \\
\hline
\end{tabular}
\caption{Readability of Layman-Style Reports. Original represents professional reports. Baseline, LLM1+P1 and LLM1+P2 indicate layman-style reports generated by different LLMs and different prompts.}
\label{tab:read-sixing}
\end{table*}

\subsection{Scaling Law}
\label{scaling}
\begin{figure*}[ht]
\small
\centering 
    \begin{subfigure}[]{0.45\textwidth}
	\includegraphics[scale=0.45]{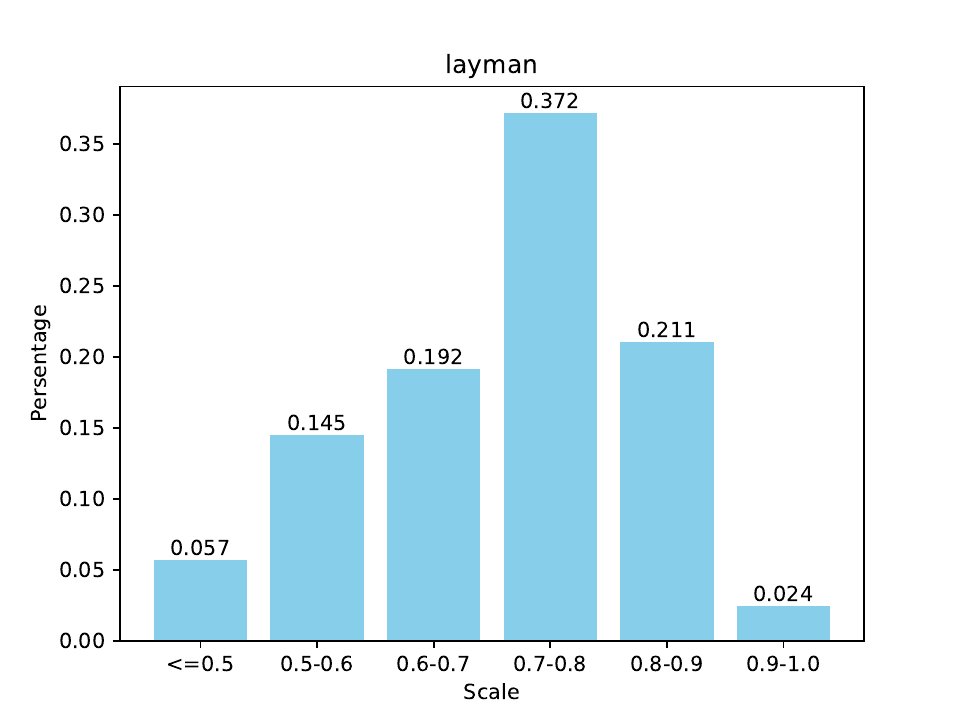}
	\caption{}
	\end{subfigure}
    \begin{subfigure}[]{0.45\textwidth}
	\includegraphics[scale=0.45]{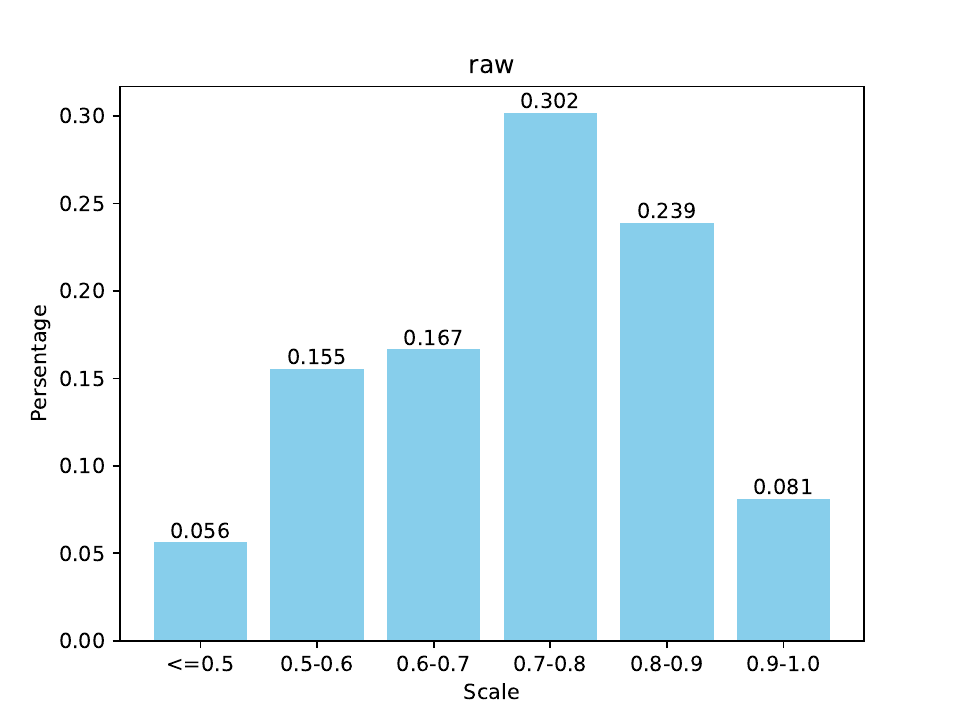}
	\caption{}
	\end{subfigure}%

	\begin{subfigure}[]{0.45\textwidth}
	\includegraphics[scale=0.45]{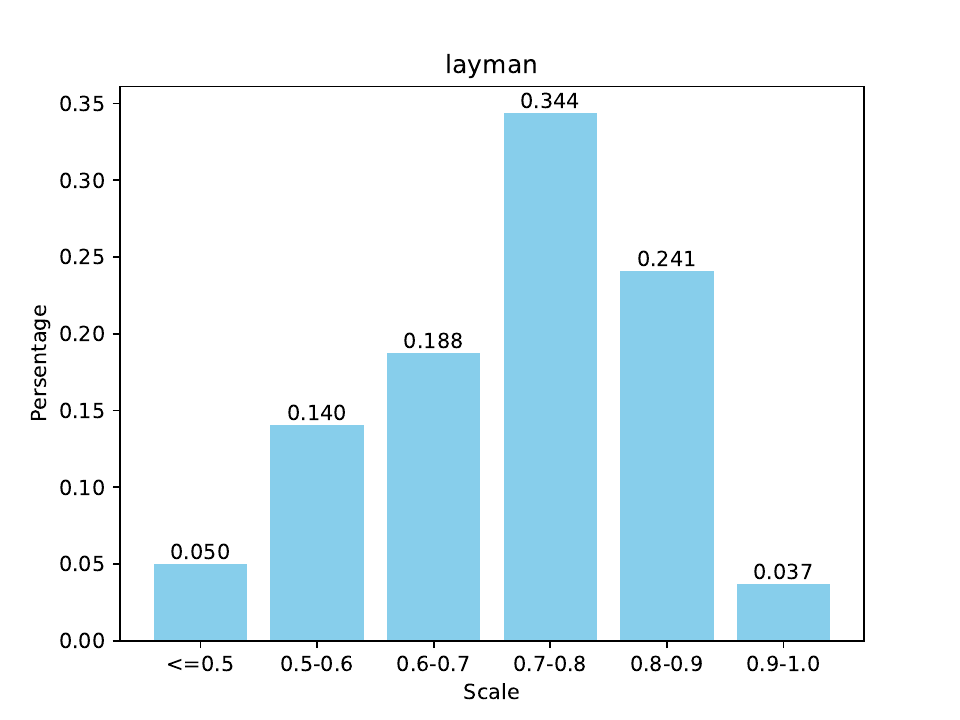}
    \caption{}
    \end{subfigure}%
    \begin{subfigure}[]{0.45\textwidth}
	\includegraphics[scale=0.45]{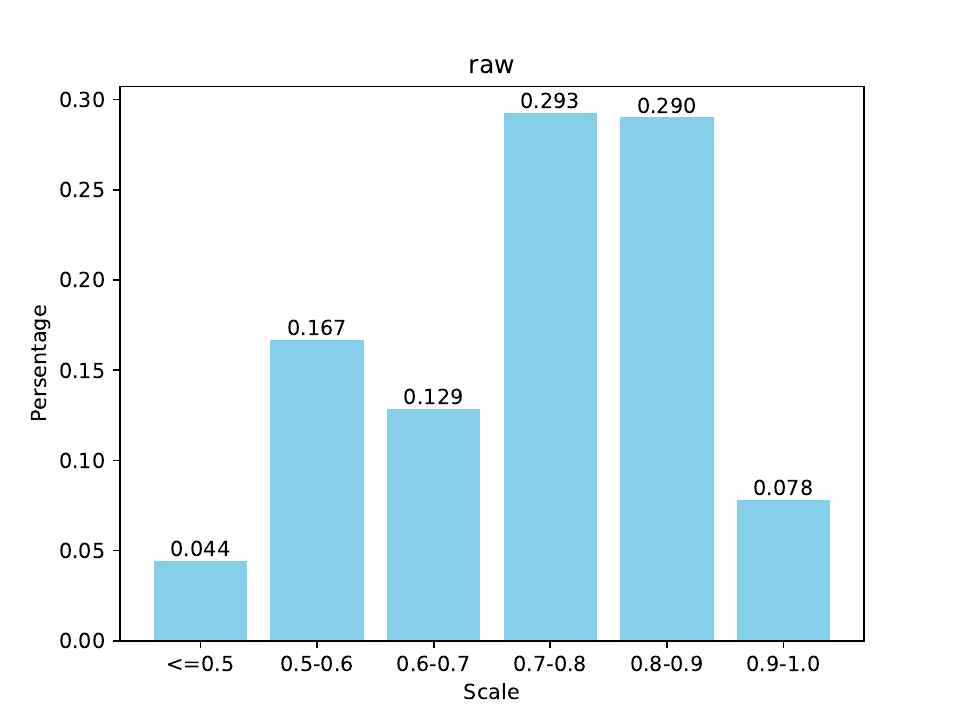}
	\caption{}
	\end{subfigure}%

    \begin{subfigure}[]{0.45\textwidth}
	\includegraphics[scale=0.45]{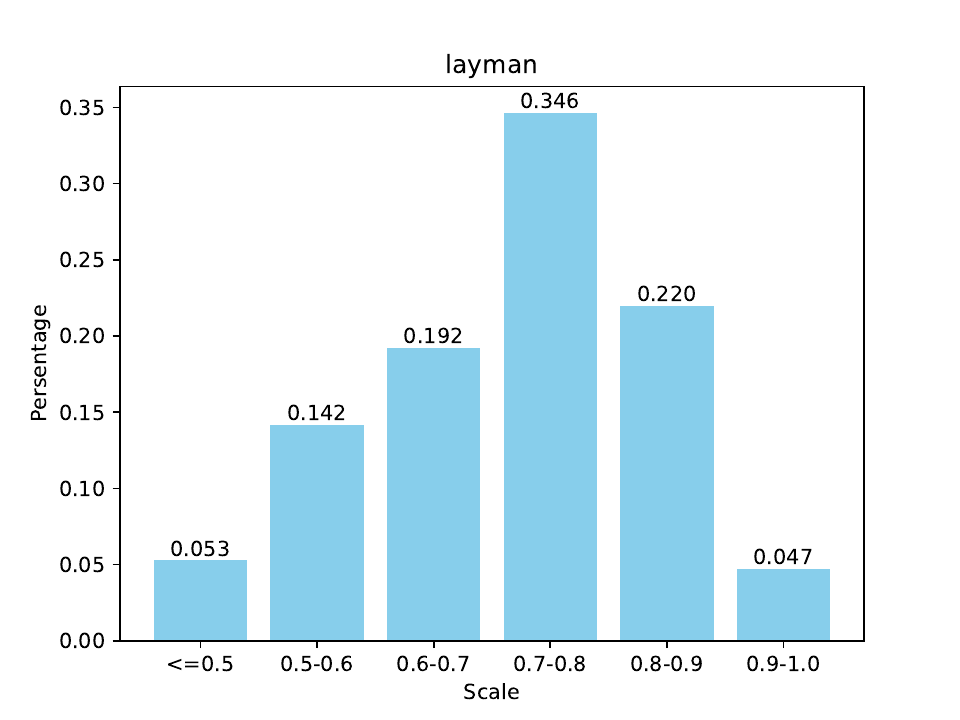}
    \caption{}
    \end{subfigure}%
    \begin{subfigure}[]{0.45\textwidth}
	\includegraphics[scale=0.45]{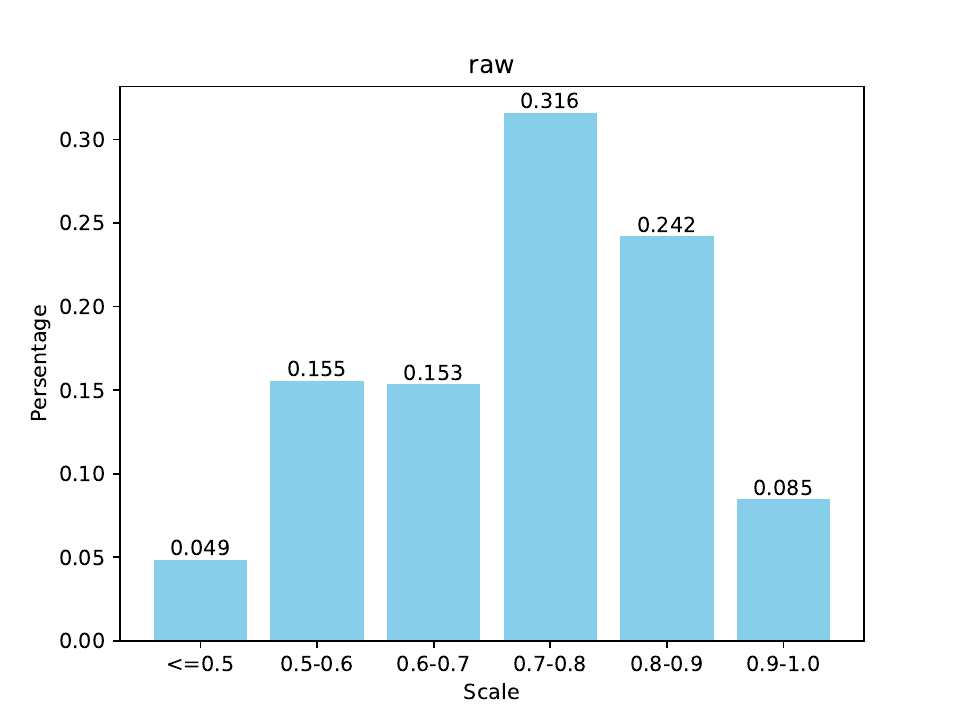}
	\caption{}
	\end{subfigure}%
 
     \begin{subfigure}[]{0.45\textwidth}
	\includegraphics[scale=0.45]{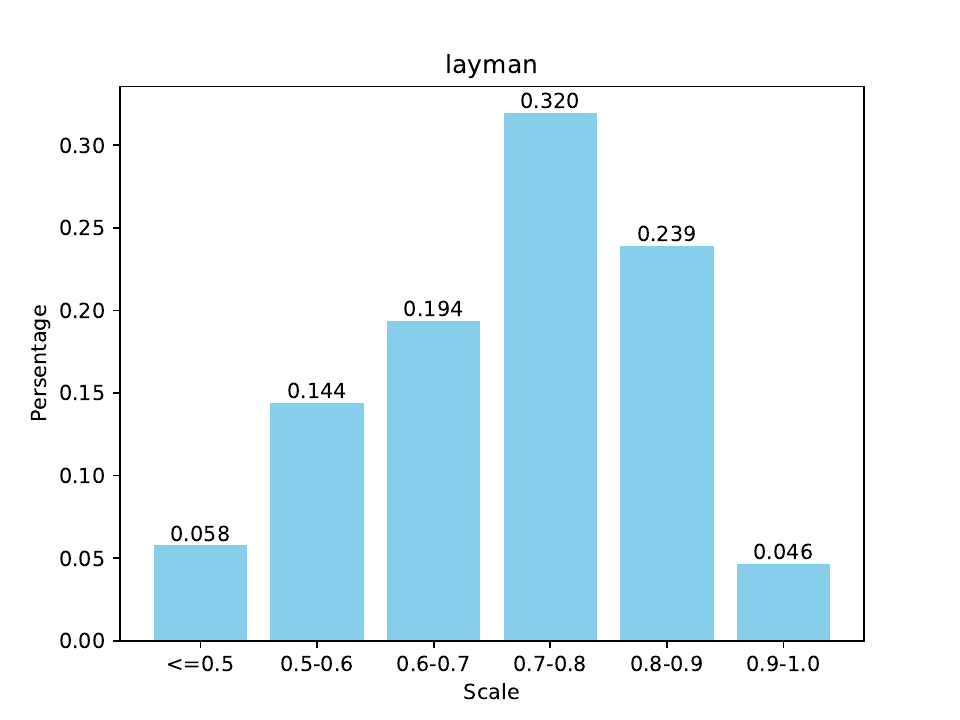}
    \caption{}
    \end{subfigure}%
    \begin{subfigure}[]{0.45\textwidth}
	\includegraphics[scale=0.45]{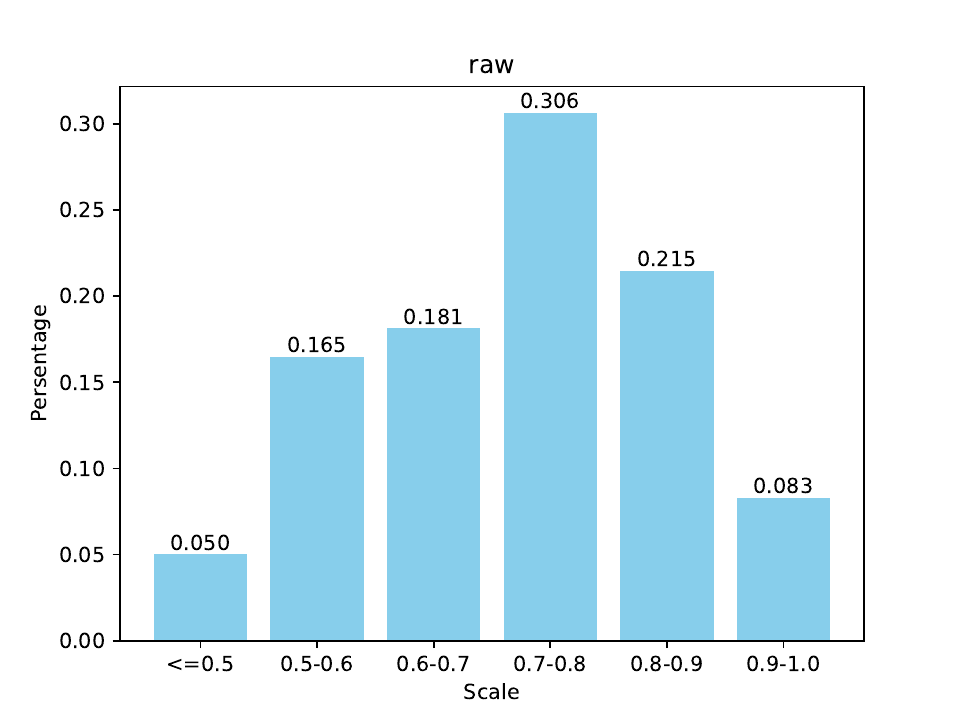}
	\caption{}
	\end{subfigure}%
\caption{Scaling law of model's semantic understanding training using report-level datasets. From up to down shows the trend for models trained by 5k, 10k, 15k and 20k respectively.}
\label{fig:label2}
\end{figure*}

As illustrated in Figure~\ref{fig:label2}, the training dataset scales are 5k, 10k, 15k, and 20k from top to bottom, respectively. We use the trained models to generate reports and calculate the semantic similarity between the generated reports and reference reports. The figures on the left represent models trained by layman's terms, while the plots on the right represent those trained using raw professional reports.

\subsection{Details of Human Annotators}
\label{human}
\noindent\textbf{Institutional Review Board (IRB).}~~Our work does not require IRB approval as it only involves semantic assessment. Our evaluation compares the semantic consistency between paragraph pairs, where the ground truth is sourced from a public dataset available on GitHub. As our task focuses solely on semantic consistency without involving any X-ray images in the evaluation process, it can be considered a common text generation task. 

\noindent\textbf{Human Annotators}~~We would like to highlight the nature of the human evaluation of this work as the assessment of semantic alignment, which makes the task fall back to the evaluation of a regular text generation task. This process is without involvement of any medical images. So we recruit human annotators from linguistic students and medical PhD students, who are professional in English reading and understanding. In addition, all of them have the right to access the MIMIC-CXR dataset.

\end{document}

%% file: Tables/concept.tex
\begin{table*}[ht]
\scriptsize
\centering
\begin{tabular}{cccl}
\toprule[2pt]
\multicolumn{4}{c}{Examples of DS \& SE} \\
\midrule[1pt]
 \multicolumn{1}{c|}{Candidate} & \multicolumn{1}{c|}{Reference}& \multicolumn{1}{c|}{Candidate layman term}& \multicolumn{1}{c}{Reference layman term}
\\ 
\midrule[1pt]
\multicolumn{1}{l|}{\begin{tabular}{p{3.2cm}} 
\sethlcolor{blue!25}
The chest x-ray shows a \hl{normal} cardiomediastinal contour and heart size.\end{tabular}}   & \multicolumn{1}{l|}{\begin{tabular}{p{3.2cm}} 
\sethlcolor{orange!40}
The chest x-ray shows low lung volumes and a \hl{mildly enlarged} heart size\end{tabular}}  & \multicolumn{1}{l|}{\begin{tabular}{p{3.2cm}} 
\sethlcolor{blue!25}
The chest x-ray shows a \hl{normal} heart and chest. \end{tabular}} & \multicolumn{1}{l}{\begin{tabular}{p{3.2cm}} 
\sethlcolor{orange!40}
The chest x-ray shows lower than normal lung volumes and a \hl{slightly enlarged} heart. \end{tabular}} 


\\   
\midrule[1pt]
\multicolumn{1}{l|}{\begin{tabular}{p{3.2cm}}
\sethlcolor{blue!25}
The chest x-ray shows well-expanded and \hl{clear lungs} without any focal consolidation, effusion or pneumothorax\end{tabular}}   & \multicolumn{1}{l|}{\begin{tabular}{p{3.2cm}} 
\sethlcolor{orange!40}
The chest x-ray shows left mid lung \hl{linear atelectasis/scarring}, without any focal consolidation or large pleural effusion\end{tabular}} & \multicolumn{1}{l|}{\begin{tabular}{p{3.2cm}} 
\sethlcolor{blue!25}
The chest x-ray shows \hl{clear lungs} without any infection, fluid, or air outside the lungs.\end{tabular}} & \multicolumn{1}{l}{\begin{tabular}{p{3.2cm}} 
\sethlcolor{orange!40}
The chest x-ray shows some minor \hl{scarring or collapse} in the left lung without any signs of localized lung infection or significant fluid.\end{tabular}} 

\\

\midrule[2pt]
\multicolumn{4}{c}{Examples of SS \& DE} \\
\midrule[1pt]
\multicolumn{1}{l|}{\begin{tabular}{p{3.2cm}}
\sethlcolor{blue!25}
Impression: \hl{No acute cardiopulmonary} process\end{tabular}}   & \multicolumn{1}{l|}{\begin{tabular}{p{3.2cm}} 
\sethlcolor{orange!40}
The impression is that there's \hl{no acute cardiac or pulmonary} process\end{tabular}}  & \multicolumn{1}{l|}{\begin{tabular}{p{3.2cm}} 
\sethlcolor{blue!25}
No serious heart or lung issues. \end{tabular}} & \multicolumn{1}{l}{\begin{tabular}{p{3.2cm}} 
\sethlcolor{orange!40}
The conclusion is no serious heart or lung issues. \end{tabular}} 

\\

\midrule[1pt]
\multicolumn{1}{l|}{\begin{tabular}{p{3.2cm}}
\sethlcolor{blue!25}
The \hl{cardiac and mediastinal} silhouettes are \hl{grossly} stable\end{tabular}}   & \multicolumn{1}{l|}{\begin{tabular}{p{3.2cm}}
\sethlcolor{orange!40}
The \hl{cardiomediastinal} silhouette \hl{appears} stable\end{tabular}} & \multicolumn{1}{l|}{\begin{tabular}{p{3.2cm}} 
\sethlcolor{blue!25}
The heart and central chest \hl{area look} stable.\end{tabular}} & \multicolumn{1}{l}{\begin{tabular}{p{3.2cm}} 
\sethlcolor{orange!40}
The heart and central chest \hl{structures appear} stable.\end{tabular}} 

\\
\midrule[1pt]
\multicolumn{1}{l|}{\begin{tabular}{p{3.2cm}} 
\sethlcolor{blue!25}
Additionally, there is no \hl{sign} of pleural effusion \hl{or} pneumothorax\end{tabular}}   & \multicolumn{1}{l|}{\begin{tabular}{p{3.2cm}} 
\sethlcolor{orange!40}
There are no pleural effusions \hl{and} pneumothorax\end{tabular}} & \multicolumn{1}{l|}{\begin{tabular}{p{3.2cm}} 
\sethlcolor{blue!25}
There are no \hl{indications} of fluid build-up or air \hl{leakage} in your lungs.\end{tabular}} & \multicolumn{1}{l}{\begin{tabular}{p{3.2cm}} 
\sethlcolor{orange!40}
There is no fluid build-up in the chest, and no air \hl{leaks} from the lungs.\end{tabular}}

\\
\bottomrule[2pt]
\end{tabular}
\caption{Samples can be categorized based on different semantics but similar expressions, as well as similar semantics but different expressions. The upperpart showcases examples of different semantics and similar expressions. Although these sentences yield a high BLEU score, they convey distinct meanings. Conversely, the lower part section presents examples of similar semantics and different expressions. Despite having a high BLEU score, these sentences express different meanings. The {\sethlcolor{blue!25}\hl{blue box}} and {\sethlcolor{orange!40}\hl{orange box}} denote the differing expressions in the reference and candidate texts.}
\label{tab:concept}
\end{table*}